# RS-CA-HSICT: A Residual and Spatial Channel Augmented CNN–Transformer Framework for Monkeypox Detection

Rashid Iqbal[1], Saddam Hussain Khan[1]*

[1]Artificial Intelligence Lab, Department of Computer Systems Engineering, University of Engineering and Applied Sciences (UEAS), Swat 19060, Pakistan

**Email:** hengrshkhan822@gmail.com

## ABSTRACT

Monkeypox (Mpox) has become a new global health threat mainly because of its rapid transmission through direct skin contact. Early and precise diagnosis of the lesion is of paramount importance for containment, while its high infectiousness and the emergence of new evolving variants further complicate public health monitoring. Therefore, this work proposes a hybrid deep learning approach, namely Residual and Spatial Learning based Channel Augmented Integrated CNN-Transformer architecture, that leverages the strengths of CNN and Transformer towards enhanced MPox detection. The proposed RS-CA-HSICT framework is composed of an HSICT block, a residual CNN module, a spatial CNN block, and a CA, which enhances the diverse feature space, detailed lesion information, and long-range dependencies. The new HSICT module first integrates an abstract representation of the stem CNN and customized ICT blocks for efficient multihead attention and structured CNN layers with homogeneous (H) and structural (S) operations. The customized ICT blocks learn global contextual interactions and local texture extraction. Additionally, H and S layers learn spatial homogeneity and fine structural details by reducing noise and modeling complex morphological variations. Moreover, inverse residual learning enhances vanishing gradient, and stage-wise resolution reduction ensures scale invariance. Furthermore, the RS-CA-HSICT framework augments the learned HSICT channels with the TL-driven Residual and Spatial CNN maps for enhanced multiscale feature space capturing global and localized structural cues, subtle texture, and contrast variations. These channels, preceding augmentation, are refined through the Channel-Fusion-and-Attention block, which preserves discriminative channels while suppressing redundant ones, thereby enabling efficient computation. Finally, the spatial attention mechanism refines pixel selection to detect subtle patterns and intra-class contrast variations in Mpox. Experimental results on both the Kaggle benchmark and a diverse MPox dataset reported classification accuracy as high as 98.30% and an F1-score of 98.13%, which outperforms the existing CNNs and ViTs. The outperformance establishes RS-CA-HSICT as a reliable, interpretable, and effective automated tool for MPox diagnosis.

## 1. Introduction:

Monkeypox (MPox) is a member of the Orthopoxvirus genus that belongs to a very close zoonotic lineage with smallpox and vaccinia viruses, showing repeated potential for outbreaks and localized epidemics [1]. Since the first identification among monkeys in 1959, this virus has been circulating among humans since its first confirmed case in 1970 [2]. This disease, while generally less lethal than COVID-19, has seen a significant surge in the number of reported cases globally [2]. The modes of transmission include direct contact with the infection from an infected person or animal and indirectly from contact with contaminated environmental surfaces. It presents symptoms mostly as fever, myalgia, and fatigue, but most characteristically in rash skin lesions that are the hallmark of this disease [3].

WHO declared MPox a Public Health Emergency of International Concern; however, MPox was a significant burden to global health in 2022. Its management includes effective isolation of the population in which the infection is taking place, strict contact tracing, and early case detection. To further consider, according to the CDC, as of January 31, 2023, there were 85,469 cases in 94 countries. No specific treatment exists; however, treatment aimed at symptoms and prevention, such as vaccination against smallpox, is very relevant [4].

Diagnosis of MPox relies on a combination of clinical evaluation and confirmatory laboratory testing. Therefore, in this work, an attempt has been made to enhance diagnostic precision and speed for MPox using emerging AI techniques in the form of ML and deep learning, specifically the Convolutional Neural Network (CNN) [5]. The value of AI in medical imaging is by now well established, with these techniques playing an increasingly key role in diagnosing and managing a wide range of diseases [6]. The most powerful among them are DL architectures, particularly CNNs and ViTs. Their strength lies in the capability for self-learning of complex, hierarchical feature representations from raw images directly [7], [8]. These models have the power to analyze medical images for discriminative features that commonly result in superior diagnostic performances compared with the classic approaches, which rely on manual feature extraction [9]. However, their actual effectiveness is normally limited by a lack of data and high computational demands. Transfer learning (TL) is usually employed to prevent over-fitting and enhance model generalization when dealing with smaller datasets.

On the other hand, the computer-aided diagnostic systems process the medical data using advanced algorithms that enable the automatic diagnosis of sicknesses [10], [11]. The rising cases of MPox infection bring out the need for effective DL-based CAD methods for image-based diagnosis. Very prominent deep learning techniques to discover hidden structural features in images include CNN and ViT. Instead of relying on convolutions, ViT relies on SA mechanisms to emphasize critical regions of an image.

Medical imaging frequently encounters challenges stemming from the disproportion between high-dimensional feature spaces and the relatively small size of available datasets, which can lead to the curse of dimensionality [6]. TL is frequently used to alleviate overfitting, particularly with small datasets [12]. Other challenges include limited availability of data, variability in infection sites and image contrast, morphological differences, and inter-class variability. The conventional ViT methods also suffer from the difficulties of local feature extraction and require a lot of computational resources.

To address these challenges, this paper introduces the Residual and Spatial Learning based Channel Augmented Integrated CNN-Transformer (RS-CA-HSICT) model, an advanced DL network that combines CNN and Transformer architectures. The model has an initial CNN block and a dual-stream network model to improve the quality of data extraction fed to the Transformer. Our hybrid approach is based on residual learning and spatial exploitation (RS) for collecting both local and global characteristics and texture variation, as in [5], [13], and improving the diagnosis performance in such conditions as MPox. Contribution: The following are some key contributions of this research study:

- The proposed RS-CA-HSICT introduces a hybrid framework that unifies transformer and CNN strengths for comprehensive MPox image analysis. RS-CA-HSICT integrates four novel components: an HSICT CNN-Transformer block, a residual CNN module, a spatial CNN block, and a Channel Augmentation (CA) that enriches the diverse feature space and enhances discrimination.
- The abstract stem CNN, custom ICT blocks with efficient multihead attention, and structured CNN layers perform homogeneous (H) and structural (S) operations. The customized ICT blocks learn global contextual interactions and local texture extraction. Additionally, H and S layers learn spatial homogeneity and fine structural details by reducing noise and modeling complex morphological variations. Moreover, inverse residual learning enhances vanishing gradient, and stage-wise resolution reduction ensures scale invariance.
- The HSICT channels are augmented with TL-based Residual and Spatial CNN blocks within the RS-CA-HSICT framework, generating a rich, diverse feature space that effectively captures subtle local texture and contrast variations. These channels are pre-processed through the new Channel-Fusion-and-Attention (CFA) module, which preserves highly discriminative global context while integrating localized structural cues and computationally efficient MPox detection.
- The spatial attention mechanism will allow for the fine selection of pixels to detect subtle patterns and minor inter-class variations in MPox images. The proposed RS-CA-HSICT framework has been tested on the publicly available Kaggle datasets and diverse datasets for MPox diagnosis and compared against benchmark state-of-the-art CNNs and the ViTs.

Section 2 reviews prior work, Section 3 outlines the proposed MPox diagnostic framework, and Section 4 details the dataset, enhancement strategies, and evaluation metrics. Section 5 reports the ablation results, and Section 6 concludes the study with directions for future research.

## 2. Literature Review

The current developments in DL have made them an integral part of the clinical world with less manual intervention. DL has shown its potential for diagnosing several diseases, including brain cancer, respiratory conditions like pneumonia and tuberculosis, and COVID-19 [14]. However, the increased prevalence of MPox, coupled with the unavailability of kits for testing, thus poses a need for alternative diagnostic methods. With a lack of experienced clinicians, care delivery at most health facilities is made even difficult. DL models would present solutions to such bottlenecks as a limited supply of kits of RT-PCR, unsatisfactory results of tests, the high cost of diagnostics, and long processing time [13]. This has led to recent research focusing on applying DL techniques to enhance triage systems and improve the performance of MPox diagnoses.

Several computer vision-based ML and DL methods have been explored. For MPox identification, Irmak et al. (2022)[15] used MobileNetV2 and VGGNet. Using MPox-infected images from the MSLD dataset, Alcalá-Rmz et al. (2023)[16] used the MiniGoogleNet model and achieved an accuracy of 97% in detecting exanthematic disorders. Ozsahin et al. (2023)[17] presented a computer-aided method for MPox detection on a dataset of digital skin images using AlexNet, VGG16, and VGG19. Bala et al. (2023)[18] designed and evaluated a tailored DenseNet-201 CNN model for MPox identification. Using DenseNet201, Sorayaie Azar et al. (2023)[19] obtained an F-score of 89.61%, 95.18% Acc, and 89.82% Sen. Khan et al. (2024)[20] used the DenseNet-201 for MPox feature extraction, training six different ML classifiers. With an accuracy of 97.55%, logistic regression exceeded the baseline DenseNet-201 model, which had a performance of 95.91%. Deb Raha et al. (2024)[21] utilized Attention-based MobileNetV2. Ahsan et al. (2024)[22] utilized the M-ResNet50 model for MPox classification. Biswas et al. (2024)[23], utilized the DarkNet53 model on the MSLD v2.0 dataset.

ViTs have been investigated recently to overcome CNN's shortcomings in MPox identification. Initially created from natural language processing (NLP), Vision Transformers (ViTs) have demonstrated efficacy in a range of vision applications, from lower- to higher-level tasks [24]. Kundu et al. (2022)[25] applied various classifiers, including ResNet50 with TL, achieving 91.00% Acc, Pre, and Sen. Their ViT model achieved 93.00% Acc, 93.00% Pre, and 92.00% Sen on the MSLD dataset. Nayak et al. (2023)[26] combined CNN with ViTB18, achieving 71.55% Acc, 49.77% Pre, 79.26% Sen, and a 61.11% F-score on the MSLD dataset. Aloraini (2023)[27] used the ViT model, achieving 94.69% Acc, 95.00% Pre, and Sen, and an F-score on a public dataset. On the MSID dataset, Arshed et al. (2024)[28] used the VIT model and obtained 93.00% Acc, Pre, recall, and F-score. Oztel (2024)[29] Using DenseNet201 with a Bagging-Ensemble ViT, and on the MSLD and PAD-UFES-20 datasets. Kundu et al. (2024)[30] utilized the model ViT-B32 with an outstanding resultant Acc rate of 97.90% on the GAN-Augmented dataset. Furthermore, Table 1 lists relevant research related to MPox detection. Despite strong performance from pre-trained CNN and ViT models, challenges remain in low-contrast feature extraction, distinction between healthy and infected cases, interpretability, generalization, and efficiency. Conventional methods capture local features poorly and struggle with computational complexity.

- CNNs primarily capture local patterns, which can weaken spatial correlations and limit their ability to model complex or large anatomical structures.
- ViTs rely on fixed-size patches, making performance sensitive to patch resolution and potentially reducing the fidelity of local feature representation.
- Deep models face vanishing gradients and high computational demands, which hinder effective training and require substantial processing resources.

The proposed framework addresses these limitations by preventing vanishing gradients, improving feature reuse, and enhancing efficiency for multi-class MPox classification.

Table 1. Previous studies employed CNN, ViT, and hybrid transformer-based models for Mpox skin disease detection.

| Author | Dataset | Model | Acc % |
| --- | --- | --- | --- |
| Ali et al. (2022) [31] | MSLD | ResNet50 | 82.96 |
| Irmak et al. (2022) [15] | MSLD | MobileNetV2 | 91.38 |
| Sahin et al. (2022) [32] | MPox Skin Lesion Dataset (MSLD). | MobileNetV2 | 91.11 |
| Altun et al. (2023) [33] | web-scraped data | hybrid MobileNetV3-s | 96.00 |
| Almufareh et al. (2023) [34] | MSID | MobileNet | 96.55 |
| | MSLD | InceptionV3 | 94.00 |
| Velu et al. (2023) [35] | MSLD | EfficientNet B3 | 96.01 |
| Sorayaie Azar et al. (2023) [19] | MPox-dataset-2022 | DenseNet201 | 95.18 |
| Bala et al., (2023) [18] | MPox Skin Images Dataset (MSID) | DenseNet201 | 93.19 |
| Nayak et al. (2023) [26] | MSLD | ResNet-18 | 99.49 |
| | MSLD | ViTB-18 | 71.55 |
| Khan and Ullah (2023) [36] | Kaggle dataset | Inception-ResNet | 97.00 |
| Eliwa et al. (2023) [37] | Mpox PATIENTS Dataset | GWO with CNNs | 95.30 |
| Alrusaini (2023) [38] | web-scraped images. | VGG16 | 96.00 |
| D. Biswas et al. (2024) [23] | MSLD v2.0 | DarkNet53 | 85.78 |
| M.M. Ahsan et al. (2024) [22] | Custom Dataset | Hybrid CNN+ViT | 89.00 |
| Raha (2024) [21] | MSID | Attention-based MobileNetV2 | 92.28 |
| G Y Oztel (2024) [29] | PAD-UFES-20 and MSLD | Bagging-Ensemble ViT with Densenet201 | 81.91 |
| Arshed et al. (2024) [28] | | ViT | 93.00 |
| Yash Suthar et al. (2025) [39] | - | Inception-based DL system | 96.71 |
| Gizachew Mulu Setegn (2025) [40] | GitHub | LGBMClassifier | 89.30 |
| Asif et al.(2025) [41] | Kaggle monkeypox dataset | JuryFusionNet CNN Ensemble | 96.8 |
| Solak, A. (2025) [42] | Custom monkeypox skin images | Hybrid Transformer + CNN | 97.4 |
| Pham et al.(2025) [43] | Monkeypox benchmark images | WDCViT Transformer Model | 96.5 |
| Islam et al. (2025) [44] | Custom multi-class skin lesion data | Multi-CNN + ML Ensemble | 97 |

## 3. Methodology

The proposed hybrid RS-CA-HSICT DL framework combines CNN and Transformer components to enhance feature representation for MPox diagnosis. The pre-processing involves applying DA as a preprocessing step to reduce bias and enhance the generalization capability of the model. The features include an HSICT CNN-Transformer block for extracting global context and local features, which is the central part of the RS-CA-HSICT model for feature extraction and classification, assisted by the feature maps from transfer-learned residual and spatial CNN modules. The feature maps obtained from these modules are first refined by the CFA block, preserving discriminative channels and suppressing redundant ones to construct a rich multi-scale feature space.

A final spatial attention further refines pixel-level selection for subtle pattern detection. In this regard, the performance of the proposed MPox detection strategy is evaluated in terms of a comparative performance analysis between: 1) the proposed RS-CA-HSICT model; 2) state-of-the-art ViTs; and 3) contemporary CNN models. The overall workflow of Mpox diagnosis is depicted in Figure 1.

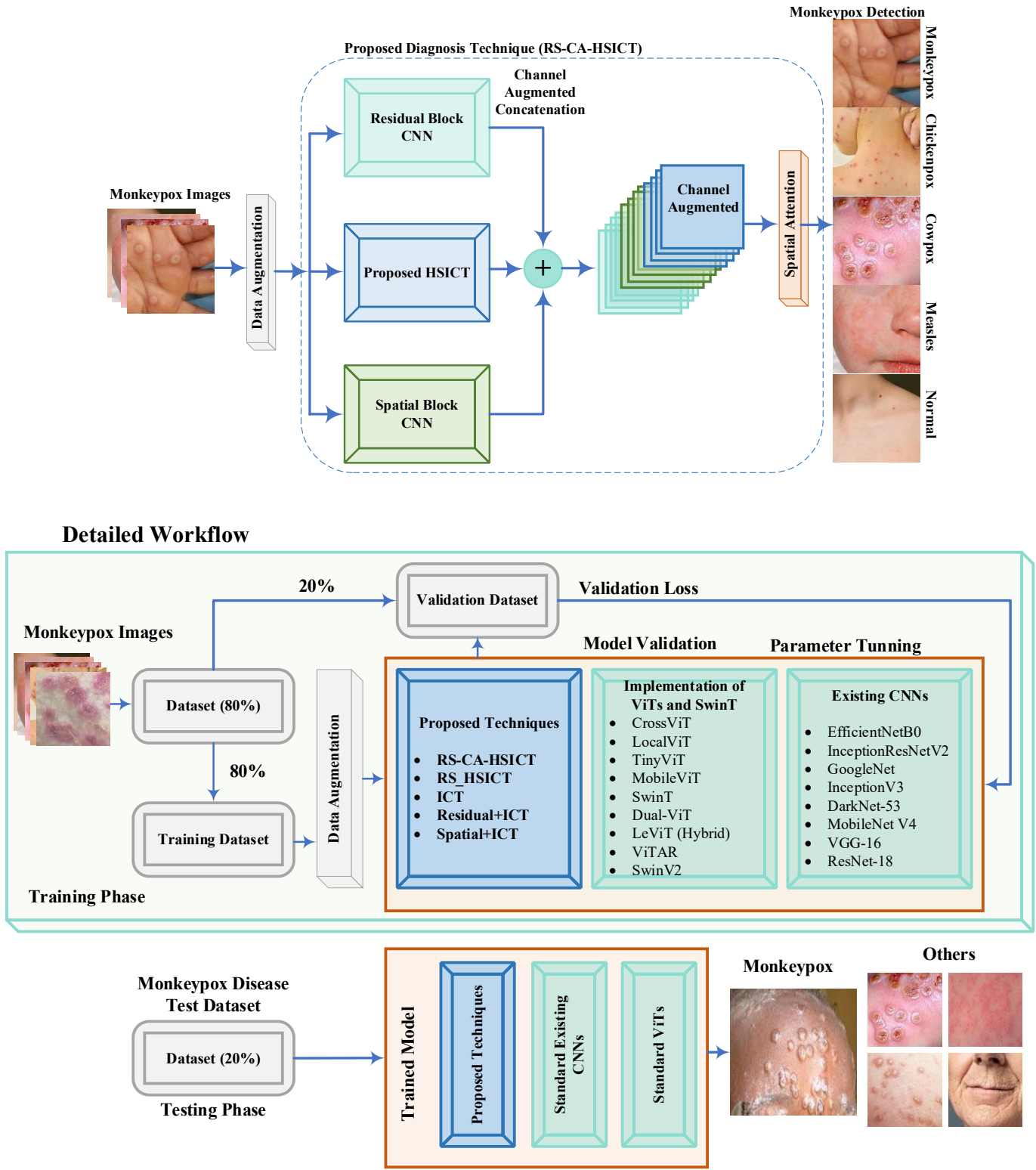

Figure 1. (A) Schematic overview and (B) detailed workflow of the proposed MPox diagnosis framework.

## 3.1. Data Augmentation

Data Augmentation (DA) plays an important role in enhancing CNN performance, especially in data-scarce scenarios. In medical imaging applications, it is sometimes very difficult to get adequate images because of privacy concerns. By training ML and DL models on a combination of original and augmented images, generalizability can be improved. Research studies employing CNNs for image-based tasks have demonstrated that DA can lead to reduced error rates.

In other words, to allow the network to identify a variety of samples found in real-world applications, the incorporation of image augmentation during training is very important. In this paper, DA techniques have been applied to the training dataset to address challenges related to class imbalance and overfitting. Usually, the MPox class suffers from class imbalance, which leads to increased false positives and thus affects the results negatively. In such scenarios, DA techniques are adopted that enhance the representation of underrepresented classes. Some methods applied in this context include random flipping (up-down, vertical, left-right, and horizontal), scaling, reflection, and shearing.

## 3.2. The proposed RS-CA-HSICT Technique

The proposed RS-CA-HSICT framework is built upon a new HSICT backbone augmented with refined feature representations from TL-based residual and spatial CNN modules. Overall, the backbone architecture consists of four sequential stages of HSICT blocks, to which the outputs of RS modules are integrated. The HSICT architecture is based its roots on hybrid CNN-Transformer models, which have been very effective in recent medical image analysis. While ViTs divide input images into discrete, non-overlapping patches that are processed by linear encoder layers, CNNs reinforce these models with their strong capabilities to represent local and structural dependencies effectively.

The architecture of HSICT enhances the efficiency of convolution, especially in the early MPox image processing steps comprising patch segmentation and tokenization. The emphasis on convolution further strengthens the extraction of salient features related to diseases. Furthermore, the residual learning module based on TL generates multiple, high-dimensional feature representations. These are combined with the output from the backbone of HSICT, thereby creating a much more robust multiscale feature space. In this, the spatial components optimize local contrast, while an integrated attention mechanism captures subtle discriminative cues and intra-class contrasts in MPox images.

To this end, the Channel-Fusion-and-Attention (CFA) block was developed to further refine the feature maps from the TL-based residual and spatial modules before augmenting them into the HSICT stream. These incoming feature maps are compressed by Global Average Pooling (GAP) to produce channel-wise attention weights, followed by a 1D convolution and fully connected layers with sigmoid activation. These weights adaptively modulate the feature responses through element-wise multiplication, preserving discriminative channels and suppressing redundant ones to allow computationally efficient fusion. Figure 2 illustrates the framework of RS-CA-HSICT and shows that HSICT, residual learning, and spatial feature exploitation are integrated. The detailed operation of all constituent modules is elaborated on in the following.

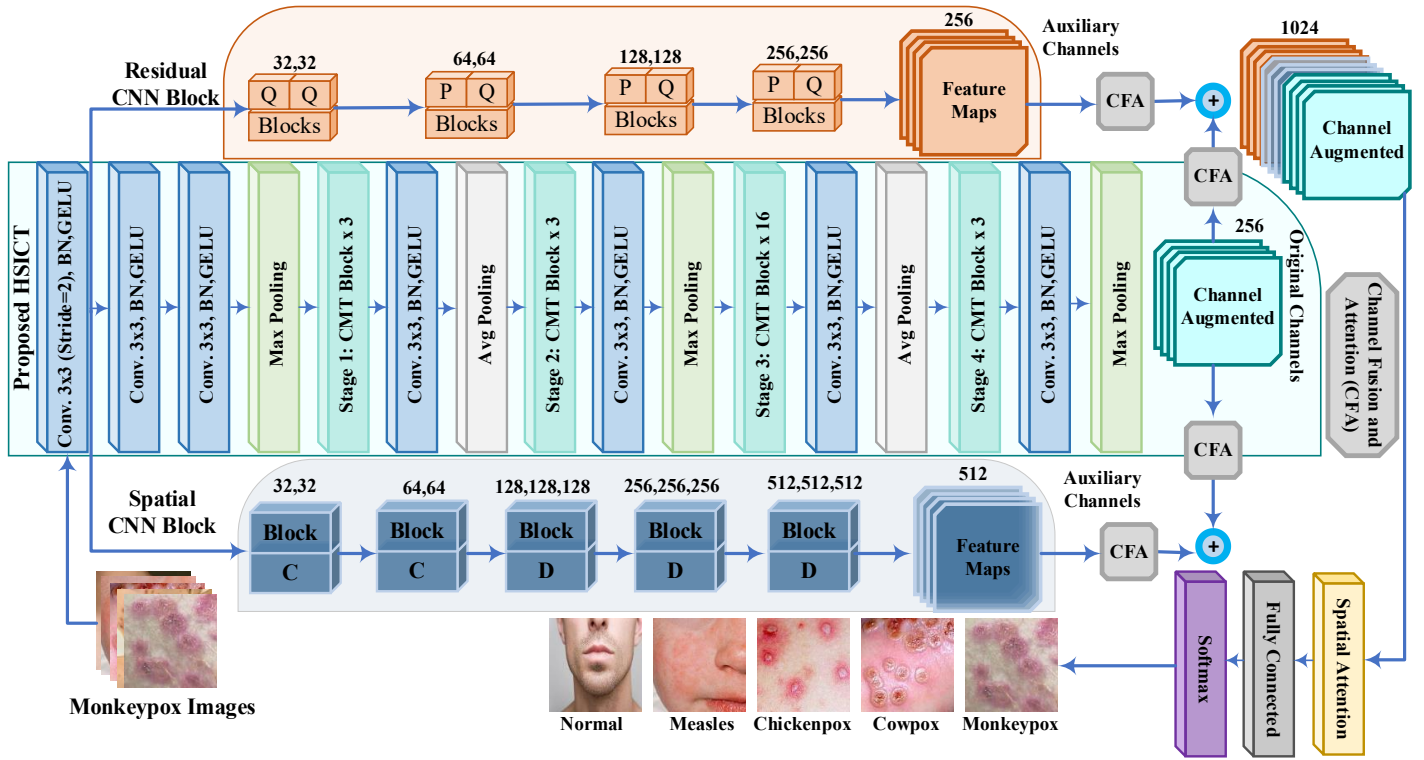

Figure 2. Detailed schematic of the RS-CA-HSICT pipeline, including stem-CNN, customized ICT, average/max-pooling, and TL-based residual and spatial CNN modules.

### 3.2.1. The Proposed HSICT Architecture

This paper concerns the design of a hybrid architecture by combining the complementary representational capabilities of CNNs and Transformer models. Based on a core HSICT block, the proposed HSICT framework integrates an abstract stem CNN with customized ICT blocks and structured CNN layers for capturing both homogeneous and structural features. First, input images are fed into an abstract CNN stem block for preliminary processing and local feature extraction Figure . In the stem CNN block, the convolutional layer $Conv_L$ is implemented using a (3x3) kernel with a stride of 2 and the output channel size of 64. Then, two more $Conv_L$ layers are attached in succession with a (3x3) kernel for further enhancement in the extraction of local feature representations.

Then, the processed features are fed into the core HSICT module. This module is based on ICT blocks incorporating an efficient variant of multi-head attention to capture global context and long-range dependencies, further enhancing interactions between features. The subsequent structured CNN layers perform homogeneous (H) operations, which ensure spatial homogeneity and noise reduction, and the structural (S) operations maintain fine structural details and model complex morphological variations. These features, after every stage, pass through a convolutional layer followed by an activation function described in Equation 1. Further, max and average pooling are applied to capture consistent regional information and fine structural details, respectively, as shown in Equations 2 and 3. While the regional operations help in noise reduction and enhance the contrast, boundary operations are effective in capturing intricate shapes and variations of the lesions. The whole process progressively reduces the resolution of the feature maps between stages and helps to make the model more invariant to minor changes for increased robustness.

Equation 1 defines 'c' as the feature map from the $\text{Conv}_L$ layer. The dimensions are 'k' and 'l'. The process of convolution involves kernels '$f$' of size 'a x b', giving an output in the range $[1 \text{ to } k - x + 1, l + y - 1]$. Equations 2 and 3 work on a window of size 'w' scanning over the convolved output $C_{k,l}$. These homogeneous (H) and structural (S) outputs are combined by using Equation 4. The weights $\gamma_1$ and $\gamma_2$ are the tuning parameters; they are tuned experimentally in order to achieve a good balance between smooth regions and well-defined edges. In this way, this combined operation smooths the local activations, hence fostering spatial invariance. Simultaneously, techniques such as edge-enhanced convolution or a Laplacian of Gaussian filter are used to capture structural transitions and edges of MPox lesions. Thus, this is basically what the homogeneous operator does: it pools regional features inside the window 'w', enhancing the local consistency.

$$C_{k,l} = \sum_{a=1}^{x} \sum_{b=1}^{y} c_{K+a-1,l+y-1} f a,b \qquad (1)$$

$$S(k,l) = C^{max}\cdot_{K,l} = max_{a=1,\ldots,w, b=1,\ldots,w} x^x k+a-1,l+b-1 \qquad (2)$$

$$H_{k,l} = C^{avg}\cdot_{K,l} = \frac{1}{w^2} \sum_{a=1}^{w} \sum_{b=1}^{w} x_{K+a-1,l+b-1} \qquad (3)$$

$$HS(k,l) = y_1 R(k,l) + y_2 B(k,l) \qquad (4)$$

The HSICT module is a four-stage consecutive block, with each stage generating multiscale feature maps that are highly effective for dense prediction tasks. Each stage incorporates the HSICT block for progressive refinement of features, as defined in Equations (5–7). This hierarchical scaling reduces the spatial resolution of intermediate features progressively, adding invariance for the sake of robustness. The HSICT block generates hierarchical feature channels at different resolutions. Finally, the four output channels from the HSICT stages are preprocessed via CFA block to achieve refined and augmented with the refined feature maps from the TL-based residual and spatial CNN modules to create the final rich multi-scale feature space for classification.

$$x_{.k,l} = \sum_{.i=1}^{m} \cdot \sum_{.j=1}^{n} \cdot x_{k+i-1,l+n-1} \cdot f_{i,j} \qquad (5)$$

$$x_{.k,l}^{max} = max_{i=1,\ldots,w, j=1,\ldots,w} x_{k+i-1,l+j-1} \qquad (6)$$

$$x_{.k,l}^{avg} = \frac{1}{w^2} \sum_{i=1}^{w} \cdot \sum_{j=1}^{w} \cdot x_{k+i-1,l+j-1} \qquad (7)$$

The $\text{Conv}_L$ extracted feature map, represented by $x_{k,l}$ is defined in Equation (5) using kernels $f_{i,j}$ where the output spans $1. x_{k+i-1,l+n-1}$. Equations 6 and 7 illustrate the application of a homogeneous along with a structural receptive field, represented by w, applied to the result of the convolution ($\mathbf{x}_{k,l}$).

**3.2.1.1. The Proposed ICT Block**

The overall architecture of ICT was proposed with four stages, each with a different design for multiscale feature channel generation in order to handle challenges effectively. In each stage, the ICT blocks are stacked in a systematic order to transform the features without hindering the spatial resolution of the input. Figure 3 gives a detailed view of the integration of blocks at each hierarchical stage. This architecture is initiated with a patch embedding layer using a (3x3) $\text{Conv}_L$ along with normalization for establishing a hierarchical representation of data in ICT. That promotes further learning of multiscale features and acts as an added advantage for MPox image analysis. Various ICT configurations with the use of strided $\text{Conv}_L$ operations enhance discriminative and scale-invariant representation.

The ICT block is designed to capture both short-range and global contextual relationships, as described in Section 3.2. In contrast to the conventional multi-head self-attention (MSA) mechanism used in ViTs, the ICT architecture incorporates a lightweight multi-head self-attention (LMHSA) module and replaces the standard multilayer perceptron (MLP) with an Inverted Residual Feed-Forward Network (IRFFN). The architecture consists of two-layer normalization sub-layers, followed in sequence by LMSA and IRFFN modules. These changes enhance the capability of the model in representing complex dependencies and enhancing feature discrimination in difficult visual tasks. Furthermore, an LPU is integrated within the ICT block to enhance the extraction of local feature capability, as shown in Figure 3. The three main components of the ICT block are mathematically represented in Equations (8–10).

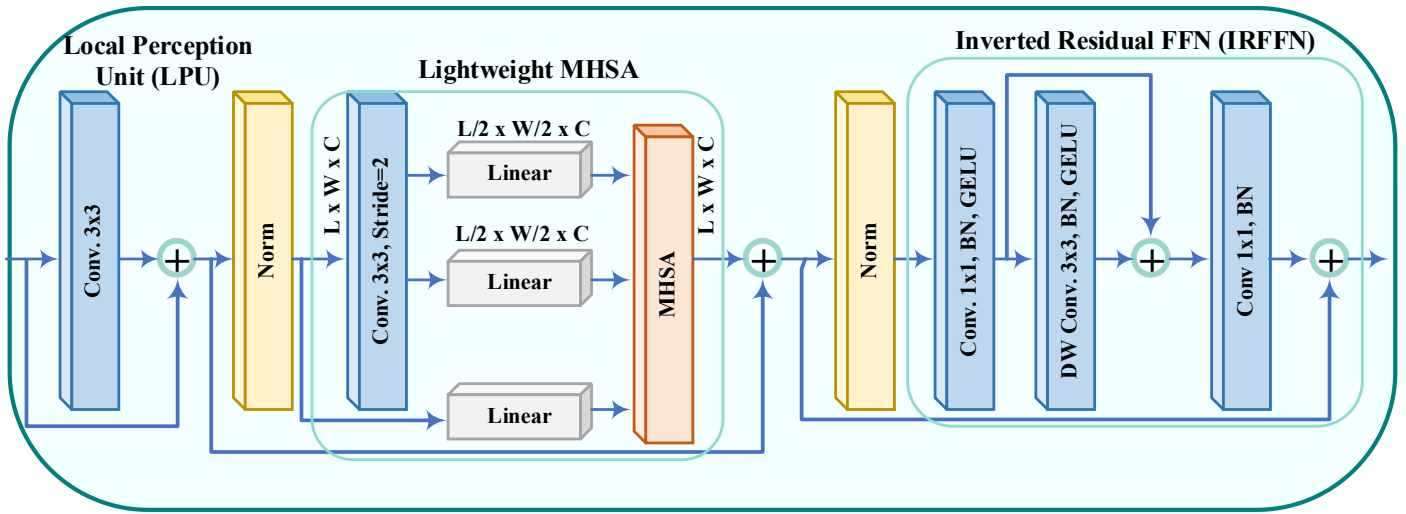

Figure 3. The proposed ICT block containing LPU, LMHSA, and IRFFN.

$$y_i = LPU.(x_{i-1}) \quad (8)$$
$$z_i = LMHSA.(LN(y_i)) + y_i \quad (9)$$
$$x_i = IRFFN.(LN(y_i)) + z_i \quad (10)$$

The output features of the $i^{th}$ block from the LPU and LMSA blocks are denoted as $y_i$ and $z_i$, respectively. To facilitate effective feature transformation and aggregation, LN is used, and several ICT modules are successively layered in every stage.

i. **Local Perception Unit**

The Local Perception Unit (LPU) is designed to address the challenge of invariance; thus, most of the work on rotation and shift augmentations has aimed to achieve translation invariance of model outputs [45]. Unlike earlier transformers that rely on fixed positional embeddings, which encode spatially differentiated positions across patches and can break invariance. LPU ensures that those issues are avoided. Quite often, ViTs fail to capture local relationships and structural information [46] within patches, which necessitates introducing the LPU. This unit extracts local features quite effectively and overcomes the drawbacks of earlier methods. The dimensions of the input 'x' are $L$, $W$, and $C$, where $L \times W$ is the size of the input and d is the feature dimension. This is represented in Equation 11 as follows:

$$LPU(x) = Conv\_L(x) + x \quad (11)$$

ii. **Light Weight Multi-Head Self-Attention Module (LMHSA)**

LMHSA block overcomes the intrinsic constraints of the single-head Self-Attention mechanism, which often focuses on a partial spatial range and thus misses salient contextual features. The SA mechanism, at the core of ViTs, is known to model explicit relationships among entities in a sequence. A few ViT variants have refined this SA module for better computational efficiency and representational capacity. Some of these employ dense global attention to enhance the connections between features, while others use sparse attention strategies to better capture long-range dependencies of features, especially in those images that are uninformative spatially. MSA extends SA by operating multiple attention heads in parallel, thereby overcoming this constraint effectively by enabling the simultaneous capture of diverse contextual relationships that improve the modeling of global interactions among visual entities.

The SA block first linearly transforms the input 'x' to obtain query (q), key (k), and value (v) matrices, and assigns them different representation subspaces. As shown in Equations 12-14, this enables MSA to learn many types of complex interactions among sequence elements:

$$A(q, k, v) = \sigma\left(\frac{q \cdot k^T}{\sqrt{d_k}}\right) \quad (12)$$

$$k' = Conv\_L.(k) \quad (13)$$

$$v' = \text{Conv\_L.}(v) \quad (14)$$

To enhance computational efficiency, a 3x3 depth-wise $\text{Conv}_L$ with stride 2 is employed to downsample the spatial dimensions of the key ($k$) and value ($v$) matrices prior to the attention operation. This allows for efficient localized information extraction. In earlier stages, convolutional blocks replace the standard transformer layers to obtain downsampled feature maps of varied spatial resolution. Consecutive tokenization of the feature maps is carried out by transformer blocks to allow hierarchical feature extraction. In addition, the attention mechanism is performed within the neighborhood of a (4x4) patch window for finer processing granularity and to incorporate more contextual sensitivity. It also includes a comparative position bias term B in the SA block, which provides the backbone of the lightweight attention mechanism, as given below:

$$\text{LA.}(q, k', v') = \sigma.\left(.\frac{qk'^T}{\sqrt{d_k}} + B\right).v' \quad (15)$$

Here, q, v, and $k^t$ denote the query, value, and inverted key matrices; $\sqrt{d_k}$ represents the scaling factor where $d_k$ is the dimension of the key matrix, and the attention mechanism, denoted by LA and σ, incorporates a non-linear activation function to enhance representational capacity.

The mechanism of MSA consists of several SA blocks with independent parameterized weight matrices in the query, key, and value subspaces. The outputs from these attention heads are summed up and linearly projected into the output space through a trainable parameter $w^0$. This operation enables ICT to adapt well to different downstream detection tasks. Specifically, in the LMHSA module, multiple attention heads independently apply the LA function to produce feature sequences of dimension $n\frac{d}{h}$. The feature sequences are then concatenated for a unified representation of size (n x d), where n stands for resolution or the number of patches, $h$ is the number of attention heads, and 'cat' means concatenation, as can be seen in Equations (16–17).

$$LMHSA(q, k', v') = \text{cat}(h_1, h_2, \ldots, h_h) \cdot w^0 \quad (16)$$

$$h_i = \text{LA}(q_i, k'_i, v'_i) \text{ Where i=1,2, \ldots, h)} \quad (17)$$

This methodical technique is ideal for intricate detection jobs because it maintains computing efficiency while improving attention capacities.

iii. **Inverted Residual Feed-Forward Network (IRFFN)**

IRFFNs represent a variant of the conventional residual block, distinguished by a restructured shortcut path and the inclusion of a convolution-based expansion layer to enhance representational capacity. A basic FFN consists of two linear layers separated by the GELU activation function that enables obtaining and integrating intricate features as represented in Equation (18). In transformers, FFN is also followed by the SA block in each encoder layer. Here, the authors have introduced 3 × 3 and pointwise, that is, 1 × 1 convolutions, $\text{Conv}_L$ inside the FFN block. As a result, the parameters and computational load effectively decrease with approximately no degradation of performance.

The IRFFN establishes a skip connection via an inverted residual block while replacing the conventional GELU activation and normalization layers with their much lighter versions. This is in order to preserve the basic principles underlying classical residual networks, which would allow smoother propagation of gradients through layers, as represented in Equations (19–20). Besides, the use of $Conv_L$ further strengthens the capability of the model in learning fine-grained local information with higher efficiency and better representational power. Equation (18) represents the non-linear activation function known as GELU, symbolized as $\sigma_g$, where $\mathbf{w}_1$ and $\mathbf{w}_2$ are weights, and $\mathbf{b}_1$ and $\mathbf{b}_2$ are biases:

$$FFN(x) = (b_2 + \mathbf{w}_2 \cdot \sigma_g(b_1 + \mathbf{w}_2 \cdot x)) \tag{18}$$

$$IRFFN.(x) = Conv.(\mathcal{F}(Conv.(x))) \tag{19}$$

$$\mathcal{F}.(x) = Conv\_L.(x) + x \tag{20}$$

These formulas describe how IRFFN uses residual connections and convolutions to increase computational efficiency while preserving strong feature extraction and integration capabilities.

### 3.2.2. CNN Residual Block

An improved stacking approach was followed in this work: The architecture incorporates a TL-based residual CNN, comprising four successive modules—denoted as (M) and (N) to progressively enhance feature learning, as depicted in Figure 4. The Point-Wise Convolution (PWC) in the block (M) makes use of a kernel size of (1x1) for inter-feature map transmission and a linear projection layer that maps the $Conv_L$ feature maps into distinct output dimensionalities. In both (M) and (N) blocks, the size of the $Conv_L$ kernel is kept as (3x3), ensuring an adequate preservation of the local receptive field described by Equations (21-22). A strategic concatenation of these blocks in association with the ICT module at the last stage allows for exploring heterogeneous feature spaces and considering the interaction among complementary features. Four successive residual blocks thus capture a wide spectrum of discriminative features, while the channel depth is increased successively from 64 to 256 for more refined feature representation and better overall learning.

The residual block employs TL for the generation of additional feature maps to increase channel diversity. These additional channels, generated from deep CNN representations through TL, are highly capable of modeling minute changes in texture and representation present in MPox images. Residual blocks are highly capable of learning fine-grained structural and contrast-based features relevant for MPox classification. This capability is further enhanced by the incorporation of the CA technique, which refines feature discriminability and enhances the prominence of diagnosis-based significant patterns.

$$y = T.(x, \{\mathbf{w}_i\}) + x \tag{21}$$

$$y = T.(x, \{w_i\}) + w_s x \qquad (22)$$

In these, the residual block allows a shortcut connection among the input 'x' and output '$y$' vectors of the considered layers, as defined by Equations 21 and 22. Here, $T(x, \{w_i\})$ signifies the residual mapping. In a two-layer model represented in Figure 4, it will be $T = w_2 \sigma(w_1 x)$. The $\sigma$ denotes the ReLU. An operation $T + x$ is enabled via a shortcut connection along with element-wise addition, followed by $\sigma(y)$. In cases of misalignment, a linear projection $w_s$ corrects the discrepancy through the shortcut connection.

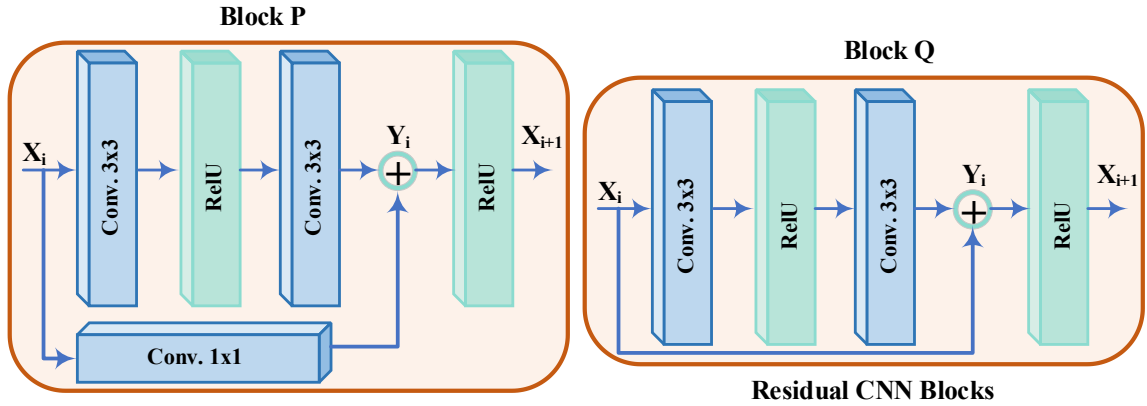

Figure 4. TL-based Residual CNN Block.

### 3.2.3. Spatial Exploitation CNN

These spatial blocks are architected to systematically extract inter-class discriminative features and preserve intra-class consistency within localized regions of MPox images. The multiple convolutional layers could be used to extract features hierarchically, therefore modeling both spatial coherence and regional homogeneity. Each spatial block employs a 3×3 $Conv_L$, batch normalization, an activation function, and refinement layers to further optimize the process of extracting locally discriminative features. Equations 23-26 and Figure 5 express the relationships among the architectural designs used. This network architecture is able to improve the capability of detecting various and discriminative features of skin lesion images, which improves the detection accuracy and promotes overall performance. Each of the spatial blocks contains the use of 3 x 3 convolutional layers, pad = 1, resulting in 2 x 2 max-pooling, stride = 2. These blocks process spatial information in successive layers through 3 x 3 filters, along with both max and average pooling methods to refine feature extraction and reinforce boundary delineation of the local structures. While convolutional kernels maintain spatial homogeneity, the information captured by the pooling layers is of high contrast and regionally important. Every time a pooling is performed, the spatial resolution reduces progressively to ensure more robustness and invariance. The network architecture is made up of five blocks in a stacked manner: convolution and pooling layers are applied recurrently to establish a hierarchy of feature extraction. This sort of configuration allows the architecture to learn complex representations deeper in the layers while retaining the necessary information in the earlier stages. The spatial blocks will lead to deeper network learning of correlated and homogeneous features, reduce the parameter count and computational load by using small compact kernels, and enhance generalization due to implicit regularizes.

$$f_l^k(p,q) = \sum_c \sum_{x,y} i_c(x,y) \cdot e_l^k(u,v) \tag{23}$$

$$N_l^k = \frac{(F_l^k) - \mu_B}{\sqrt{\sigma_l^k + \varepsilon}} \tag{24}$$

$$T_l^k = g_a(F_l^k) \tag{25}$$

$$Z_l^k = g_p(F_l^k) \tag{26}$$

For both blocks (P) and (Q), the size of the $Conv_L$ is 3 x 3; this defines an effective local receptive field, represented in Equation 23. In each convolutional layer, different neurons are assigned distinct convolutional kernels, which divide the input image into small receptive regions, hence enabling them to do feature extraction locally and provide detailed spatial patterns with fine resolution. The element $i_c(x,y)$ represents a pixel located at spatial coordinates $(x,y)$ embedded in the input image tensor $l_c$, while $e_l^k$ denotes the spatial position of the $k^{th}$ convolutional kernel within the $l^{th}$ layer. The result of the $k^{th}$ convolutional operation in layer $l$ is expressed as a feature map $F_l^k = [f_{.l}^k(1,1) \ldots, f_{.l}^k(p,q)\ldots, (f_{.l}^k(P,Q)]$ where each element corresponds to a spatial activation. Batch normalization is applied to reduce internal feature map variations. Batch normalization of a feature map($F_{.l}^k$) to a normalized feature map ($N_l^k$) is given by Equation 24, where, $\mu_B$ and $\sigma_{.B}^2$ are the mean and variance across a mini-batch, respectively, and $\epsilon$ is a constant added for numerical stability. An activation function introduces the nonlinear operation, as in Equation 25, where $(F_l^k)$ is the output after applying the nonlinear activation function $g_a(.)$. Downsampling or pooling summarizes the data from the receptive field by extracting the highest response, as in Equation 26, where $Z_l^k$ is the pooled feature map from the $l^{th}$ layer connected to the $k^{th}$ input feature map $(F_{.l}^k)$ and, $g_p(.)$ represents the pooling process.

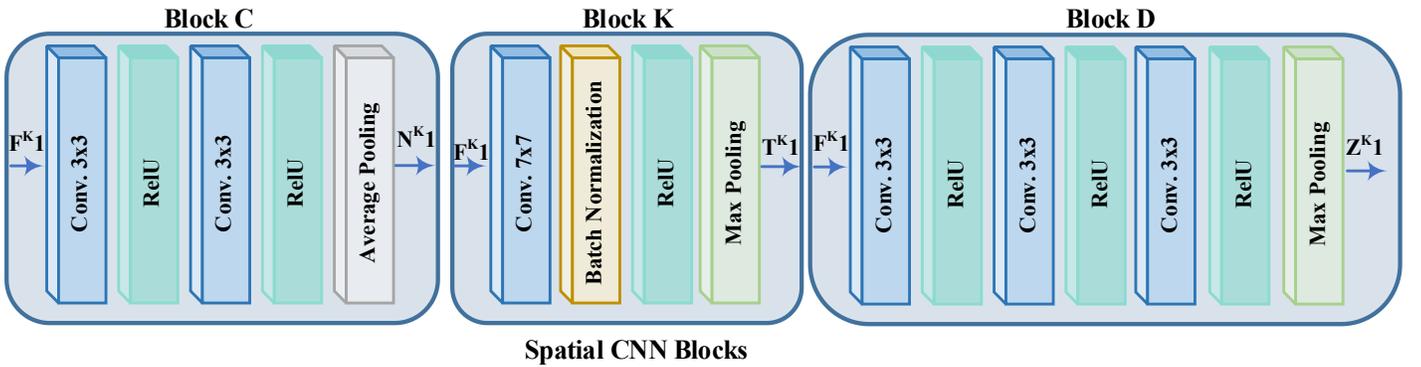

Figure 5. TL-based Spatial CNN Block Designing.

### 3.2.4. Channel-Fusion-and-Attention Block

The new CFA block is employed in the proposed framework to ensure that only the most relevant features are augmented to the main network. The CFA block refines the feature maps generated by the TL-based residual and spatial CNN modules before they are fused with the HSICT stream, preserving discriminative channels and suppressing redundant ones for efficient computation. The process in the CFA block goes as follows: GAP is applied to shrink each input feature map into one representative value, effectively compressing the spatial dimensions. Then, one-dimensional convolution and a fully connected layer are used to process the resulting representations into a channel-wise attention map via sigmoid activation (Figure 6). This attention map adaptively enhances the feature maps by element-wise multiplication and residual addition, summarized by Equations 27-30. The mechanism of channel attention adaptively allocates importance weights to different feature channels, enhancing discriminative responses while suppressing redundant information.

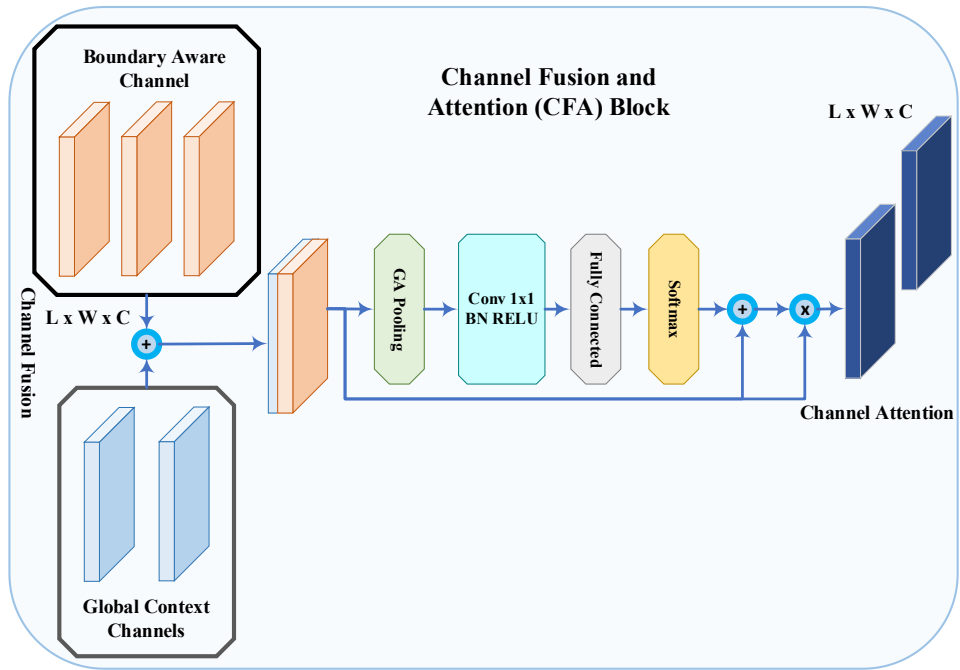

Figure 6. Architecture of the Channel-Fusion-and-Attention (CFA) block.

Here, GAP, Conv, and FC are short for global average pooling, convolution, and fully connected layers, respectively. $\oplus$ and $\otimes$ indicate element-wise addition and multiplication, respectively, as shown in Figure 6 and the referenced equations.

$$C_{Fusion} = b(X_{RS} \| C_{HICT}) \in R^{(C_B+C_G)k \times l} \qquad (27)$$

$$C_{RF} = SM(FC(Conv.(GAP(C_{Fusion})))) \qquad (28)$$

$$CA_i = SM(FC(CRF))) \qquad (29)$$

$$CA_{Final} = C_i + C_i \otimes CA_i \qquad (30)$$

### 3.2.5. Channel Augmentation

CA is adopted to enhance the learning capability of the proposed HSICT model by fusing it with refined feature maps from TL-based residual and spatial modules. The auxiliary learners are developed to recognize different and complex local variations and textures. CA refines the feature maps from the residual and spatial modules through the Channel-Fusion-and-Attention block. The refinement output from the CFA block augments the feature channels of the backbone HSICT. The integration creates a rich multi-scale feature space where the HSICT channels capture global and structural context while the augmented channels contribute enhanced local detail. This merged representation is outstanding in capturing the texture variations that are necessary for MPox detection. This augmentation is equivalent to channel-wise concatenation, formally given by Equation 31, and can be seen in Figures 1 and 2.

$$CA\ x_{HSICT} x_{Residual} x_{Spatial} = Concat(CFA x_{Residual} CFA x_{spatial}, x_{HSICT}) \quad (31)$$

Specifically, in the Equation, $x_{HSICT}$ represents the feature maps from the main HSICT backbone while $CFA(x_{Residual} x_{spatial})$ are the refined feature maps from the TL-based modules after processing with the CFA block. Concat denotes channel-wise concatenation to increase the dimensionality of the feature space.

### 3.2.6. Spatial Attention-Based Approaches

Spatial attention will then be conducted on the boosted feature maps for further enhancing spatial discrimination, refining feature selection at the level of a pixel, as shown in Figure 7. This will allow the network to emphasize selectively the diagnostically relevant regions and suppress less informative areas, a factor critical in detecting subtle patterns and intra-class contrast variations of MPox lesions. The presented mechanism will produce a 2D spatial attention map highlighting the informative areas. It does this, as described in Equations 32–34, by compressing the input feature map $\mathbf{x}_{Boosted}$ along the channel dimension by both average and max pooling to collect the spatial cues. The resulting feature maps are then joined together and subjected to convolution with the intent of getting a spatial attention map $W_{spatial}$ with values between 0 and 1 due to sigmoid activation.

$$\mathbf{x}_{SA\_out} = \mathbf{W}_{pixel} \cdot \mathbf{x}_{Boosted} \quad (32)$$

$$x_{relu} = \sigma_1(\mathbf{W}_x \mathbf{x}_{Boosted} + \mathbf{W}_{SA} SA_{m,n} + b_{SA}) \quad (33)$$

$$\mathbf{W}_{pixel} = \sigma_2(f(x_{relu}) + b_f) \quad (34)$$

$$x = \sum_{a=1}^{A} \sum_{b=1}^{B} v_a\ \mathbf{x}_{SA\_out} \quad (35)$$

$$\sigma(x) = \frac{e^{x_i}}{\sum_{c=1}^{C} e^{x_c}} \quad (36)$$

This attention map is applied to the input features through element-wise multiplication to generate the final refined output $x_{SA\_out}$, amplifying salient spatial features. Here, $\otimes$ is element-wise multiplication, $\sigma$ refers to the sigmoid function, and Conv represents a convolutional layer defined in Equations (35-36). The output $x_{SA\_out}$ contains spatially refined features where relevant regions for Mpox detection are enhanced. In Equation (32), $\mathbf{x}_{Boosted}$ refers to the input feature map, and $W_{.pixel}$ is the weighted pixel within the scale [0, 1]. The output $x_{SA\_out}$ depicts the highlighting of relevant regions while reducing the effects of irrelevant features. Equations (33) and (34) explain activation $\sigma_1$ and $\sigma_2$, biases $b_{PA}$ and $b_f$, and transformation $W_x$, $W_{PA}$, and function f. Equations (35) and (36) conclude with the number of neurons and softmax activation, parameterized by $V_a$ and $\sigma$.

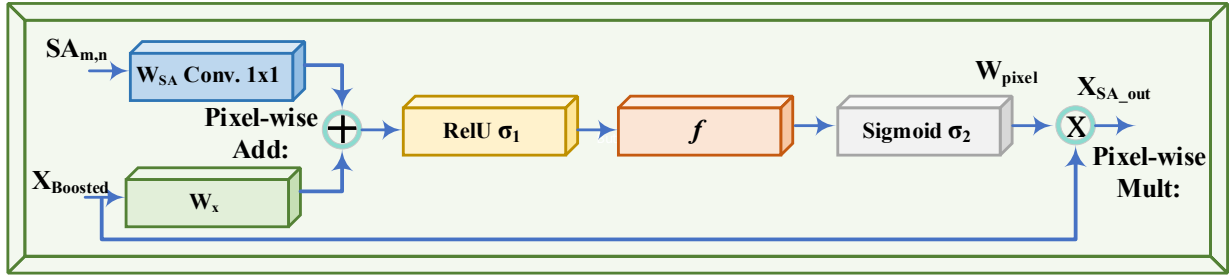

Figure 7. Spatial Attention Block.

### 3.3. Implementation of CNN and ViT Techniques

A comprehensive comparative analysis was performed in order to rigorously evaluate the performance of the proposed RS-CA-HSICT model against several state-of-the-art deep learning architectures. The studied benchmark consisted of three sets of models, namely: 1) CNNs: EfficientNetB0, InceptionResNetV2, GoogleNet, InceptionV3, DarkNet53, MobileNet V4, VGG16, and ResNet18, which are recognized by their strong capabilities for local feature extraction; 2) ViTs: CrossViT, LocalViT, TinyViT, MobileViT, and ViTAR, which represent the state-of-the-art approaches for modeling global dependencies; 3) Hybrid Architectures: LeViT, DualViT, SwinT, and SwinV2 have demonstrated promising performances in an attempt to combine the best of both worlds, namely CNNs and Transformers [47], [48]. These models have been selected due to their state-of-the-art performance in image classification problems and are representative of a wide variety of architectural philosophies, thus providing a very sound basis to evaluate critically the contributions of our newly proposed RS-CA-HSICT framework.

### 4. Experimental Setup:
### 4.1. Dataset Details

This research exclusively used the MPox dataset, verified and annotated by experts, and publicly accessible sources [49]. The MPox Kaggle dataset [49] has been used for this research, which contains images from five classes: MPox, Chickenpox, Measles, Cowpox, and Normal. The MPox dataset is publicly available on Kaggle, which includes comprehensive data, including MRI scans of more than 16000 subjects. This MPox dataset allows for efficient training and testing of DL models, leading to improved MPox detection and helping to further tune diagnostic algorithms and treatment protocols for MPox infection in the face of its increasing global burden. Due to its manageable size and diversity, and imbalance of classes, it invites more researchers for further research and refinement in diagnosing and treating MPox. Sample images of each class are included in Figure 8, while Table 2 describes the dataset distribution.

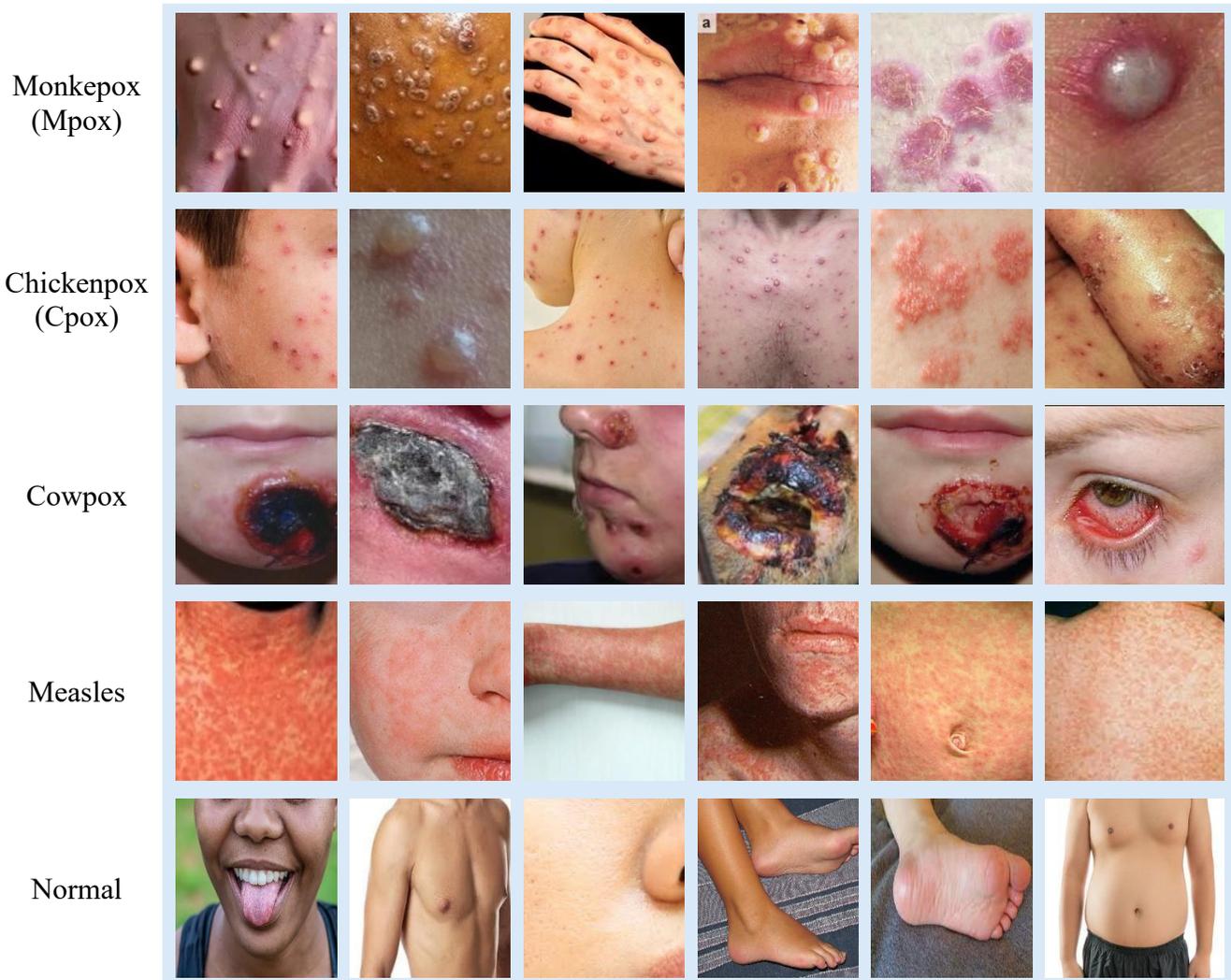

Figure 8. The Dataset comprised of MPox, Chickenpox, Measles, Cowpox, and Normal.

Table 2. Benchmarked Mpox and other skin lesion disease details.

| Characteristics | Overview |
| --- | --- |
| Total | 16630 Samples |
| Measles | 1540 Samples |
| Chickenpox | 2100 Samples |
| Cowpox | 1850 Samples |
| Monkeypox | 7950 Samples |
| Normal | 3190 Samples |
| Train, Validation (80%) | (10643, 2,661) |
| Test (20%) | (3,326) |
| Image Size | 224 x 224 x 3 |

### 4.2. Experimental Setup and Model Evaluation

The proposed RS-CA-HSICT model, together with the comparative ViT and CNN architectures, has been trained by using the Adam optimization algorithm with a primary learning rate of $10^{-3}$. The learning rate is decayed by

85% every 20 epochs, while enforcing a constant regularization strength of 0.04 via weight decay during the entire training process. In this regard, the cross-entropy loss function was considered to quantify the classification error so that it could deal with the class imbalance problem. Training was performed on a batch size of 16 examples with a dropout of 0.3 at the output layer to prevent overfitting and enhance model generalization. Experiments have been conducted using Python, TensorFlow, and Jupyter Notebook on a system featuring an Intel Core i9-12th Generation processor, 64 GB DDR4 RAM, 1 TB NVMe SSD, 3 TB HDD, and an NVIDIA GeForce RTX 4070 Ti GPU.

This dataset, obtained from Kaggle, was preprocessed and then assessed using hold-out cross-validation, with 20% of the data consistently allocated for validation without overlap across runs to facilitate robust evaluation. The results on these validation sets were then combined to determine overall model performance. Model performance was computed based on different metrics, namely F1-score, Acc, Pre, Sen, ROC/PR curves, and the corresponding AUC according to Equations 37-41. To enhance diagnostic sensitivity for MPox, efforts were directed toward increasing the rate of true positive (TP) while minimizing the false negatives (FN), The Standard Error (S.E.) for sensitivity, as defined in Equation 37, facilitates the estimation of its 95% Confidence Interval (CI) [7]. A z-value of 1.96 was employed to compute the S.E. for constructing the 95% CI.

$$Acc = \frac{TP+TN}{Total} \times 100 \qquad (37)$$

$$Sen = \frac{TP}{TP+FN} \times 100 \qquad (38)$$

$$Pre = \frac{TP}{TP+FP} \times 100 \qquad (39)$$

$$F-score = 2 \times \frac{Pre+Sen}{Pre+Sen} \qquad (40)$$

## 5. Results and Discussion

This section provides a thorough evaluation of the experimental results. This paper also includes ablation studies to confirm the effectiveness of the proposed HSICT and RS-CA-HSICT architectures. The RS-CA-HSICT architecture's performance is compared to state-of-the-art models like ViTs, SwinT, CNNs, and hybrid combinations of ViT-CNNs on the Kaggle dataset and diverse datasets. The results based on the classification evaluation metrics, such as Acc, Pre, Sen, F1-score, and AUC, are presented in detail in Table 3 and Figure -12. In this regard, the proposed RS-CA-HSICT model was used for multi-class classification, where images are classified into five distinct categories, namely MPox, Chickenpox, Cowpox, Measles, and Normal. It achieved the best performance and outperformed the existing state-of-the-art results of ViTs and CNNs with 98.0% Acc. The RS-CA-HSICT model thus showed a considerable improvement compared to the baseline HSICT, which only reported 96.45% Acc, hence justifying the effectiveness of the added components. The final model achieved a PRAUC of 0.9875 and an ROC-AUC of 0.9893. These reflect that the CA technique is effectively applied, and the three-stream structure consisting of HSICT, residual learning, and spatial exploitation is effective in feature encoding and generalization. Figure 9 Confusion matrix shows that most misclassifications occurred between classes, such as Cowpox and Chickenpox, which implies difficulty for the model to identify these visually similar classes.

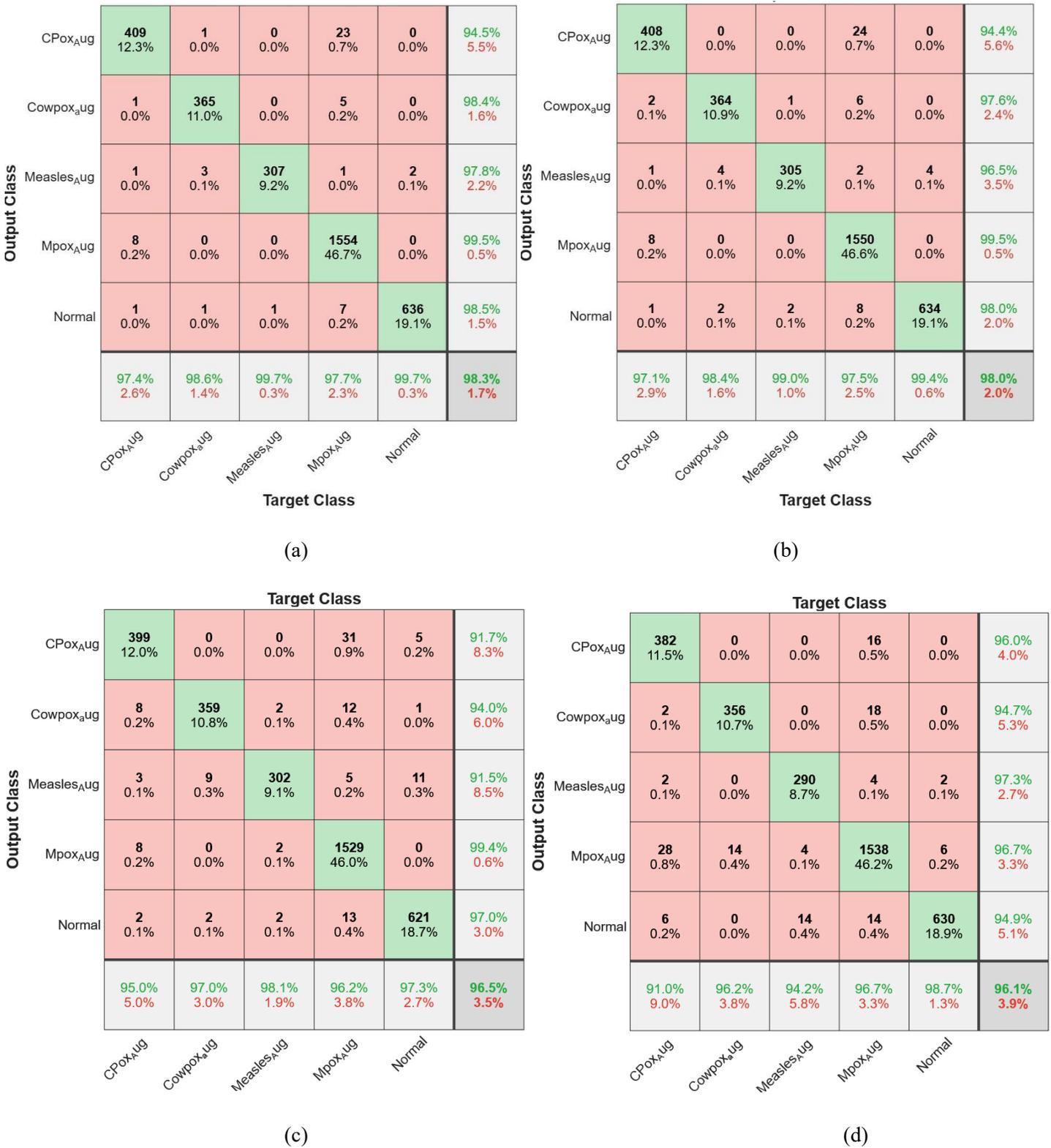

Figure 9. Confusion matrix for the evaluated techniques: (a) the proposed RS-CA-HSICT (CFA), (b) the proposed RS-CA-HSICT, (c) the customized ICT, and (d) the LeViT (Hybrid) model.

Table 3. Performance of the proposed implementation setup.

| Model | Acc. % | Sen. % | Pre. % | F1-score % | FLOPs | Inf.T (ms) | Tim/Ep. |
|---|---|---|---|---|---|---|---|
| EfficientNetB0 | 85.81 | 82.42 | 86.78 | 84.55 | 1.39 | 5.4 | 2.32 |
| InceptionResNetV2 | 86.17 | 88.02 | 82.14 | 84.98 | 13.1 | 12.3 | 4.05 |
| GoogleNet | 90.08 | 89.90 | 87.45 | 88.66 | 1.5 | 6.1 | 3.39 |
| InceptionV3 | 91.46 | 90.95 | 90.18 | 90.56 | 5.7 | 10.3 | 3.55 |
| MobileNet V4 | 92.18 | 91.17 | 91.67 | 91.42 | 10.7 | 11.5 | 2.95 |
| DarkNet-53 | 93.21 | 92.90 | 91.61 | 92.25 | 12.6 | 14.8 | 3.28 |
| VGG-16 | 94.41 | 94.07 | 93.17 | 93.62 | 4.3 | 6.1 | 2.42 |
| ResNet-18 | 94.77 | 94.06 | 94.28 | 94.17 | 3.6 | 7.9 | 2.85 |
| Existing ViTs | | | | | | | |
| CrossViT | 92.96 | 91.37 | 92.42 | 91.89 | 4.5 | 11.2 | 3.65 |
| LocalViT | 93.21 | 92.90 | 91.61 | 92.25 | 5.0 | 12.0 | 3.68 |
| TinyViT | 94.41 | 94.07 | 93.17 | 93.62 | 6.0 | 12.5 | 3.70 |
| MobileViT | 95.13 | 94.19 | 94.34 | 94.27 | 4.6 | 10.5 | 2.92 |
| SwinT | 95.31 | 94.99 | 94.93 | 94.96 | 8.0 | 14.3 | 3.75 |
| Dual-ViT | 96.03 | 95.02 | 95.82 | 95.42 | 4.1 | 9.5 | 3.58 |
| LeViT (Hybrid) | 96.09 | 95.36 | 95.92 | 95.64 | 9.5 | 15.2 | 3.08 |
| SwinV2 | 96.21 | 95.21 | 96.04 | 95.62 | 10.5 | 16.8 | 2.95 |
| Proposed Setup | | | | | | | |
| ICT | 96.45 | 95.75 | 96.46 | 96.10 | 12.2 | 18.0 | 2.85 |
| Spatial (S)+CA-ICT | 97.14 | 96.57 | 97.34 | 96.95 | 14.0 | 19.5 | 3.12 |
| Residual (RS)+CA-ICT | 97.53 | 96.97 | 97.55 | 97.26 | 16.0 | 21.5 | 3.22 |
| RS_CA-ICT | 97.84 | 98.03 | 96.83 | 97.43 | 17.0 | 22.0 | 4.07 |
| RS-CA-HSICT (PA) | 98.00 | 97.72 | 97.9 | 97.81 | 17.5 | 22.5 | 4.12 |
| RS-CA-HSICT (CFA) | 98.30 | 98.05 | 98.22 | 98.13 | 16.8 | 21.5 | 3.26 |

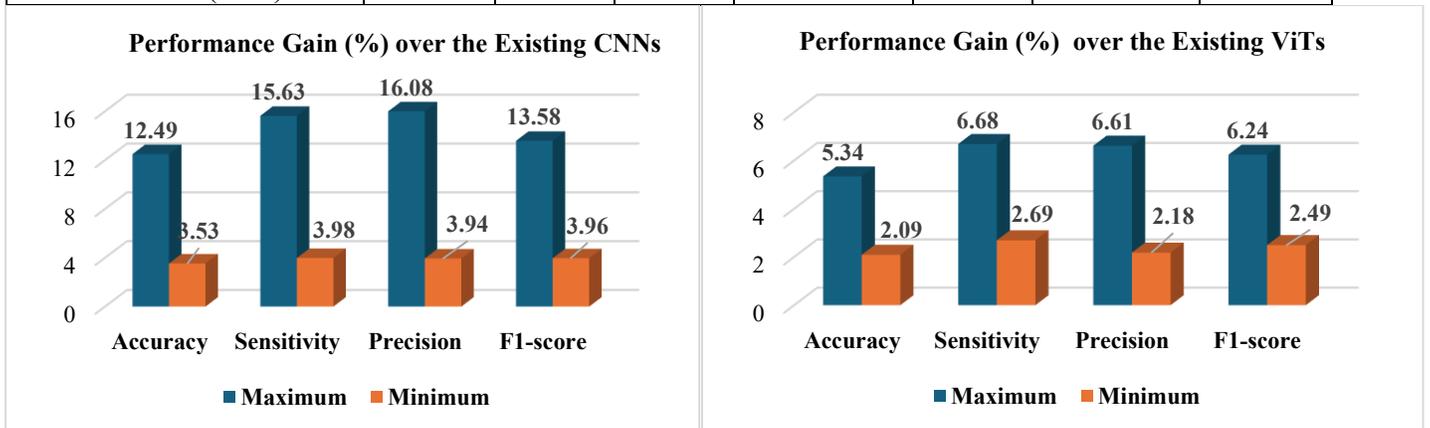

Figure 10. Illustrates the comparative performance gains achieved by the proposed RS-CA-HSICT model over conventional CNN, ViT, and SwinT architectures.

For the MPox class, an analysis of its confusion matrix showed that 98.3% MPox cases were correctly identified, or there was a 1.7% misclassification rate. Furthermore, the complexity of the skip-connection-enabled HSICT architecture was explored to determine how efficient it was. The RS-CA-HSICT methodology had lower training complexity compared to traditional CNNs, ViTs, and SwinT, as evident from Table 3. The model required a small computational load and therefore reduced the training time compared to the other models from the ViT family. Therefore, it obtained high performance with a streamlined architecture and correspondingly allowed faster convergence. However, in a few of the other hybrid CNN-ViT models, oscillations in their performance could be observed during convergence toward the optimal solution.

## 5.1. Performance Evaluation with Existing ViTs and CNNs

This section presents a detailed comparative analysis of the proposed RS-CA-HSICT framework for MPox detection with other leading models tested on the Kaggle and diverse datasets. In this work, several CNN, ViT, and SwinT models, including EfficientNetB0, InceptionResNetV2, GoogleNet, InceptionV3, DarkNet-53, MobileNet V4, VGG-16, ResNet-18, CrossViT, LocalViT, TinyViT, MobileViT, SwinT, Dual-ViT, LeViT (Hybrid), ViTAR, and SwinV2, have been employed in the diagnosis of MPox. In light of this, Table 3 presents the multi-class performance of the proposed model, demonstrating superior results relative to conventional DL methods on the same dataset. It attained significant improvements by a margin ranging between 3.59% to 12.19% in Acc over traditional CNNs designed for local feature learning, as illustrated in Figure 10.

Moreover, compared with existing ViTs that stress the learning of global features, the RS-CA-HSICT obtained performance gains ranging from 2.87% to 3.23%, while gains over SwinT were from 1.72% to 2.69% (Illustrated in Figure 10), with a significant rise of 3% to 11.8% in PR-AUC. A comparative performance test against the latest approaches on the Kaggle MPox dataset is presented in Table 3. The results depicted that SwinT V2 models tend to be much better in performance than traditional CNN and ViT models, especially for Acc, which reached as high as 96.28%, Sen equaled 95.22%, Pre equaled 96.09%, F1-score was 95.62%, and PR-AUC has been 3%. Though the hybrid models, namely LeViT and SwinT V2, achieved competitive results, the proposed RS-CA-HSICT consistently obtained the highest scores on all the important metrics, which further verifies that its unique design synergistically fuses HSICT, residual, and spatial features with channel augmentation.

## 5.2. Systematic Ablation Analysis of Architectural Components

The proposed RS-CA-HSICT was comparatively evaluated with a wide range of state-of-the-art models, including convolutional, transformer-based, and hybrid deep learning models, using the Kaggle and diverse MPox dataset, as shown in Table 3. Common CNN architectures include EfficientNetB0, InceptionResNetV2, GoogleNet, InceptionV3, DarkNet-53, MobileNet-V4, VGG-16, and ResNet-18; these are powerfully inductive with strong capabilities in local feature extraction but limited by their small global receptive fields. Thus, even the best performance among these CNNs was only 94.77% by ResNet-18. On the other hand, ViT's and hierarchical attention models, such as CrossViT, LocalViT, TinyViT, MobileViT, ViTAR, and SwinV2, showed better long-range dependency and global contextual cues capturing capabilities but at higher computational costs, with generally weaker sensitivities to subtle local textures, which are crucial for the accurate characterization of MPox lesions. Indeed, even the strongest baseline within this category, SwinV2, achieved a maximum accuracy of only 96.21%, suggesting continued limitations in achieving an appropriate balance within local-global representational fidelity.

Table 4. Comparative analysis with the previous study.

| Model | Accuracy | Sensitivity | Precision | F1-Score |
|---|---|---|---|---|
| **Existing CNNs** | | | | |
| MobileNetv2 [32] | 91.11 | 90.00 | 90.00 | 90.00 |
| ResNet50 [31] | 82.96 | 83.00 | 87.00 | 84.00 |
| ShuffleNet-V2 [50] | 79.00 | 58.00 | 79.00 | 67.00 |
| DarkNet53 [23] | 85.78 | 82.46 | 86.92 | 84.20 |
| DenseNet201 [19] | 95.18 | 89.82 | - | 89.61 |
| MobileNetV2 [15] | 95.4 | 94.2 | 93.8 | 94 |
| **Existing ViTs** | | | | |
| ViT [25] | 93.00 | 91.00 | 93.00 | 92.00 |
| ViTB18 [26] | 71.55 | 79.26 | 49.77 | 61.11 |
| ViT [27] | 94.69 | 95.00 | 95.00 | 95.00 |
| ViT [28] | 93.00 | 93.00 | 93.00 | 93.00 |
| **Hybrid Techniques** | | | | |
| Xception-CBAM-Dense [51] | 83.90 | 89.10 | 90.70 | 90.10 |
| VGG-16 naïve Bayes [52] | 91.11 | - | - | - |
| 13 DL models. Ensemble approach [53] | 87.13 | 85.47 | 85.44 | 85.4 |
| Various ML and DL models [54] | 98.48 | 89.00 | 91.11 | - |
| RestNet50 with TL [25] | 91.00 | 90.00 | 91.00 | 90.00 |
| CNNs with GWO algorithm [37] | 95.30 | 98.10 | 95.60 | 96.80 |
| GoogleNet and Metaheuristic Optimization [55] | 94.35 | 95.00 | - | 92.00 |
| optimized hybrid MobileNetV3 [33] | 96.00 | 97.00 | - | 98.00 |
| Ensemble ViT with Densenet201 [29] | 81.91 | 74.14 | 87.17 | 78.16 |

The RS-CA-HSICT was designed to address these very shortcomings through its novel incorporation of a hybrid CNN-Transformer backbone with auxiliary residual and spatial feature streams, derived via TL and further refined through a Channel-Fusion-and-Attention mechanism. This allows for the joint exploitation of local texture patterns, structural homogeneity, and global semantic dependencies within a unified multi-scale representation. Empirical results support such a design, where the proposed model posts a state-of-the-art accuracy of 98.30% corresponding to absolute gains of 3.53% over the strongest CNN baseline and 2.09% over the best-performing ViT/ SwinT variant. Furthermore, this model consistently outperforms all competing architectures on a host of additional metrics, including Sensitivity, Precision, F1-score, and PR-AUC. These findings affirm the robustness of the RS-CA-HSICT framework while also underlining its superior capability to capture the complex local-to-global variations characteristic of MPox lesions. Moreover, the proposed model outperforms leading DL methods, including SwinT, ViT, and standard CNNs, as demonstrated by assessment metrics, ROC-PR curves, and in Principal Component Analysis (PCA)-based analysis Tables 3 and 4, and Figures 8-12).

### 5.3. PR/ROC Curve

The diagnostic rate curves evaluate the discriminative capacity of the RS-CA-HSICT model across numerous threshold settings, evaluating its generalization amongst moderate and other classes. PR/ROC curves are used to analyze the prediction accuracy for the MPox classes, distinguishing between MPox, Chickenpox, Cowpox,

Measles, and Normal (Figure 11). These curves are generated by comparing the predicted probabilities against the ground-truth labels, and a summary of model performance is shown in Tables 3-4 with 20% label availability. Excellent performance is observed by the RS-CA-HSICT model, with a PR-AUC of 0.9875 and an AUC-ROC of 0.9893; the proposed model surpasses all other active learning approaches. In addition, this model shows a PR-AUC increase ranging from 3% to 11.8% and a 1.84% AUC-ROC increase. The finest ROC curves and utmost AUC scores for the compared CNN, ViT, and SwinT are from RS-CA-HSICT, according to Figure 11. The deep blue ROC curve, with data points clustering near the top-left corner, reflects superior classification performance, while the green PR curve indicates strong precision-recall characteristics with points concentrated toward the top-right region.

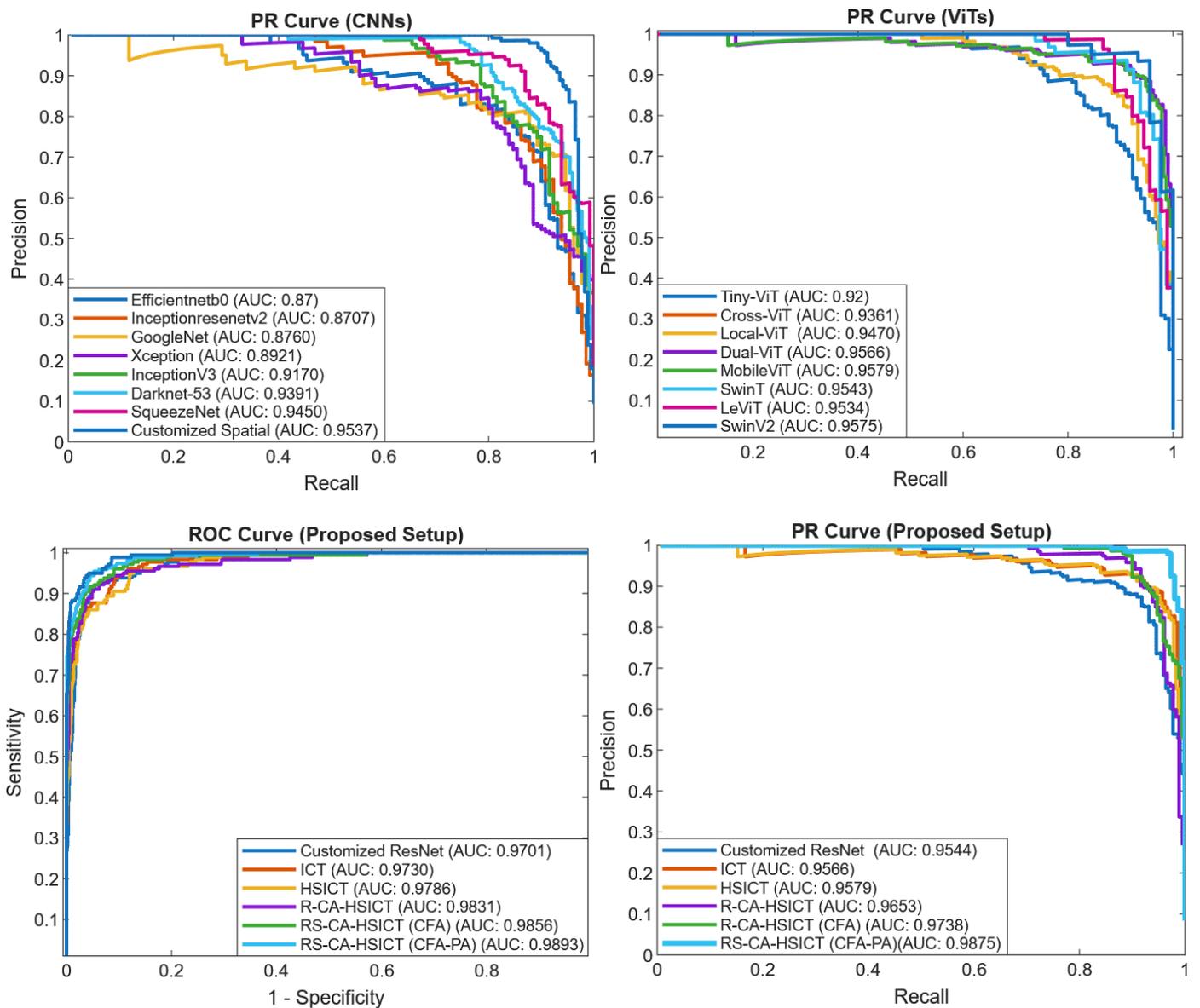

Figure 11. Detection rate analysis of the proposed (RA-CA-HSICT) and existing CNNs/ViTs.

## 5.4. Feature Space Visualization

To validate the proposed framework's effectiveness, feature maps were extracted from each layer of the RS-CA-HSICT using an unseen test set. However, the high dimensionality of these features complicates direct visualization. As part of the proposed pipeline, pooling layers and FC layers are subsequently followed by PCA for dimensionality reduction, projecting the extracted features into a 2D space. PCA is selected for its ability to preserve local feature geometry more effectively, enhancing the quality of clustering and visualization due to its nonlinear properties. The PCA was employed to visualize the embedded features extracted from the penultimate fully connected layer across four distinct models: RS-CA-HSICT, HSICT, and LeViT, utilizing the Kaggle dataset.

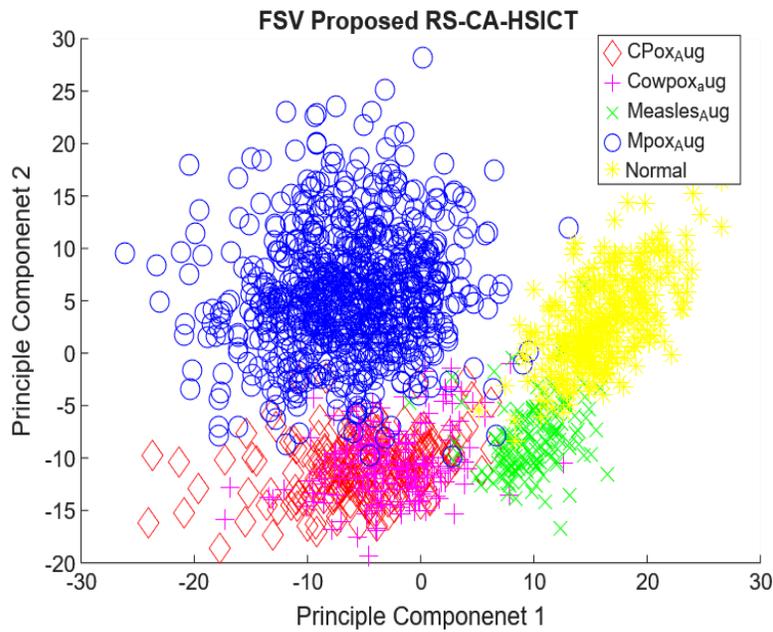

(a)

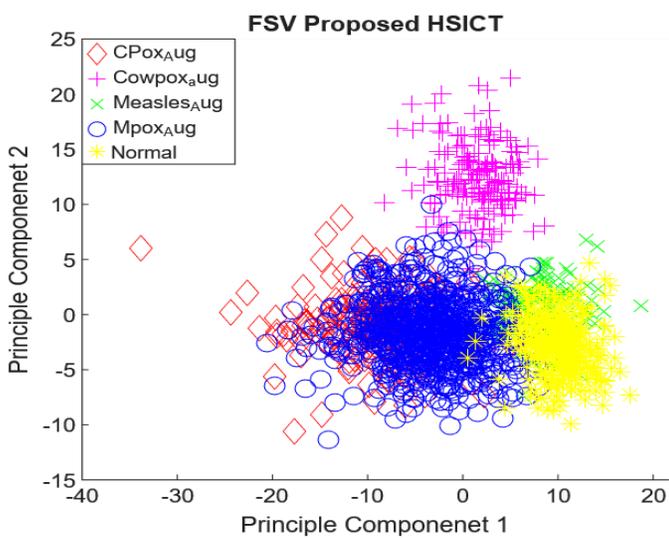

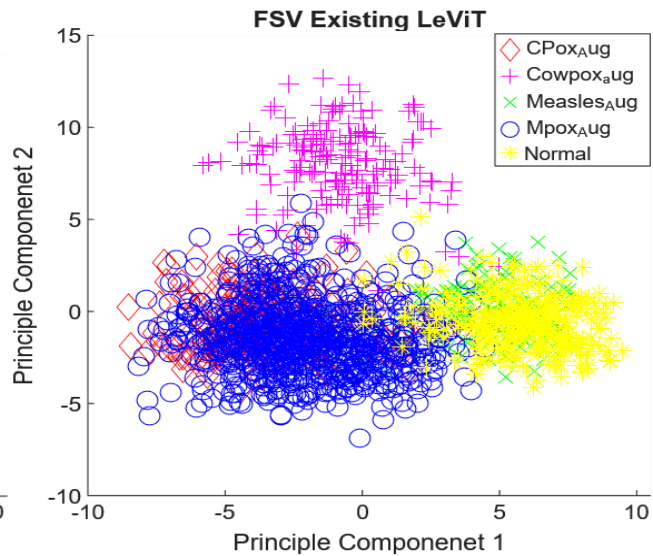

(a)                                                                                      (c)

Figure 12. presents the PCA-based feature embeddings for: (a) the proposed HSICT with CA, (b) HSICT without CA, and (c) the LeViT baseline.

Figure 12 highlights multiple PCA segments, each demonstrating that the proposed RS-CA-HSICT model can extract more clearly separable features related to the MPox, Chickenpox, Cowpox, Measles, and Normal conditions. This further shows that the proposed hybrid HSICT-based contrastive learning framework enhances the discriminative power of the features. Figure 12 depicts some of the PCA embeddings on: (a) the proposed HSICT with CA, (b) HSICT without CA, and (c) the existing LeViT. The various infectious data are shown in red, blue, magenta, and green points, respectively. The PCA results reveal a higher separability between PC1 versus PC2 of the proposed RS-CA-HSICT, pointing toward enhanced feature discrimination and higher classification accuracy, compared with other methods.

## 6. Conclusion

This work presented the RS-CA-HSICT framework to overcome persistent limitations in MPox image classification, such as poor feature representation in low-contrast images, a high computation burden, and instability due to deep architectures. The method combines structured convolutional operations, residual learning, spatial exploitation, and an enhanced CNN Transformer hybridization strategy to generate a coherent, multi-scale representation of MPox lesions. Hierarchical improvements from ICT to S-CA-ICT to R-CA-ICT and finally to RS-CA HSICT consistently show systematic gains in all metrics. The proposed RS-CA-HSICT framework with CFA further enhanced accuracy, sensitivity, precision, and F1-score to 98.30%, 98.05%, 98.22%, and 98.13%, respectively, surpassing powerful CNNs and ViTs, including LeViT and SwinV2. A compact HSICT stem and lightweight transformer blocks minimize redundancy and preserve stable gradient propagation, while the H and S operations preserve fine morphological cues related to MPox discrimination. The CA module enriches the feature space without extra computational depth, and the CFA block selectively refines high-value channels, hence reducing FLOPs while retaining discriminative strength. All of these combine to yield a more efficient learning pipeline compared to other comparable hybrid architectures that are bound to incur higher inference times due to their extensive self-attention operations. More importantly, the incorporated spatial attention module conducts pixel-level refinements that allow the model to detect subtle inter-class variations in a highly diverse MPox manifestation. Finally, RS-CA-HSICT yields a reliable diagnostic framework that offers superior performance and interpretable feature representations. Future work may extend this architecture to broader domains of biomedical imaging where low-contrast lesions, heterogeneous patterns, or scarce annotated data are persistent features.

## Abbreviation

| Full Form | Acronym | Full Form | Acronym |
|---|---|---|---|
| Monkeypox | MPox | Computer-aided diagnostic | CAD |
| Deep Learning | DL | Local Perception Unit | LPU |
| Convolutional Neural Network | CNN | Light-Weight Multi-Head Self-Attention | LMHSA |
| Vision Transformer | ViT | Accuracy | Acc |
| Channel Augmented | CA | Precision | Pre |
| Sensitivity | Sen/Recall | Principal Component Analysis | PCA |
| World Health Organization | WHO | True positive | TP |
| False negative | FN | Confidence Interval | CI |
| Standard Error | S.E | Pixel attention | PA |
| Transfer learning | TL | Inverted-Residual Feed-Forward-Network | IRFFN |
| Feed-Forward Network | FFN | Self-Attention | SA |
| Multi-head self-attention | MSA | Multilayer perceptron | MLP |
| Data augmentation | DA | MPox skin lesion dataset | MSLD |
| MPox skin image dataset | MSID | Residual and Spatial Learning | RS |
| Integrated CNN-Transformer | ICT | lightweight attention | LA |
| Point-Wise Convolution | PWC | Machine Learning | ML |
| Natural Language Processing | NLP | Channel-Fusion-and-Attention | CFA |
| Global Average Pooling | GAP | Artificial Intelligence | AI |
| Homogenous | H | Structural | S |